\documentclass{article}

\usepackage{bm}
\usepackage{array,url,amssymb,bbding,cite}
\usepackage{spconf,amsmath,graphicx,setspace,epstopdf,booktabs}
\usepackage{tabularx, makecell, multirow, multicol, arydshln, stfloats}
\usepackage{color}

\title{IMPROVING END-TO-END CONTEXTUAL SPEECH RECOGNITION WITH FINE-GRAINED CONTEXTUAL KNOWLEDGE SELECTION}

\name{Minglun Han$^{1,2,3,*}$, Linhao Dong$^{3}$, Zhenlin Liang$^{3}$, Meng Cai$^{3}$, Shiyu Zhou$^{1}$, Zejun Ma$^{3}$, Bo Xu$^{1,2}$\thanks{* This work was done when the first author was an intern with Bytedance AI Lab. This work was supported by the National Key R\&D Program of China under Grant No.2020AAA0108600 and Strategic Priority Research Program of the Chinese Academy of Sciences under Grant No.XDA27030300.}}
\address{
$^{1}$Institute of Automation, Chinese Academy of Sciences\\
$^{2}$School of Artificial Intelligence, University of Chinese Academy of Sciences\\
$^{3}$Bytedance AI Lab\\
\small \tt \{hanminglun2018, zhoushiyu2013, xubo\}@ia.ac.cn, \\
\small \tt \{donglinhao, liangzhenlin.lzl, caimeng.1, mazejun\}@bytedance.com}

\begin{document}
\ninept
\renewcommand{\baselinestretch}{0.8812} \normalsize

\maketitle

\begin{abstract}
Nowadays, most methods for end-to-end contextual speech recognition bias the recognition process towards contextual knowledge. Since all-neural contextual biasing methods rely on phrase-level contextual modeling and attention-based relevance modeling, they may suffer from the confusion between similar context-specific phrases, which hurts predictions at the token level. In this work, we focus on mitigating confusion problems with fine-grained contextual knowledge selection (FineCoS). In FineCoS, we introduce fine-grained knowledge to reduce the uncertainty of token predictions. Specifically, we first apply phrase selection to narrow the range of phrase candidates, and then conduct token attention on the tokens in the selected phrase candidates. Moreover, we re-normalize the attention weights of most relevant phrases in inference to obtain more focused phrase-level contextual representations, and inject position information to help model better discriminate phrases or tokens. On LibriSpeech and an in-house 160,000-hour dataset, we explore the proposed methods based on an all-neural biasing method, collaborative decoding (ColDec). The proposed methods further bring at most 6.1\% relative word error rate reduction on LibriSpeech and 16.4\% relative character error rate reduction on the in-house dataset.
\end{abstract}

\begin{keywords}
Contextual speech recognition, contextual biasing, collaborative decoding, knowledge selection
\end{keywords}

\section{Introduction}
\label{sec:introduction}

In recent years, many end-to-end (E2E) automatic speech recognition (ASR) approaches, such as connectionist temporal classification (CTC) \cite{graves2006connectionist, graves2014towards}, recurrent neural network transducer (RNN-T) \cite{graves2012sequence}, attention-based encoder-decoder (AED) \cite{ChorowskiBSCB15, chan2016listen, DBLP:conf/icassp/BahdanauCSBB16, dong2018speech}, have been widely employed in voice assistants, online meetings, etc. However, recognizing context-specific phrases in these scenarios remains to be improved because most contextual contents are rare in training data. For instance, the contextual contents for voice assistants are usually contacts, song playlists, etc., and the contextual contents in meetings are usually the names of those who attend meetings and some technical terms. Injecting contextual knowledge to bias decoding process of these E2E ASR models has become an important research field.

Currently, the most widely known contextual biasing methods for different E2E models are shallow fusion \cite{aleksic2015bringing, hall2015composition, williams2018contextual, HeSPMAZRKWPLBSL19, zhao2019shallow}, attention-based deep context \cite{pundak2018deep, alon2019contextual, chen2019joint, bruguier2019phoebe} and trie-based deep biasing \cite{le2021deep, le2021contextualized}. Among these methods, shallow fusion fuses a finite state transducer (FST) compiled from a list of biasing phrases into the decoding process but does not add any neural networks. In contrast, the all-neural attention-based biasing method \cite{pundak2018deep} and its extensions on other models \cite{jain2020contextual} generally encode phrases with an encoder and integrate the relevant context with an attention module at each time step. Though the all-neural attention-based methods outperform shallow fusion \cite{pundak2018deep}, it still suffers from some problems, such as the high correlation between a large number of phrases \cite{pundak2018deep} and the confusion between similar biasing phrases \cite{alon2019contextual, chen2019joint, bruguier2019phoebe}. In \cite{pundak2018deep}, the performance degradation when injecting lots of distractors reveals that the high correlation between hundreds of biasing phrases tends to prevent the model from distinguishing target phrases from other similar ones. Meanwhile, in \cite{alon2019contextual, chen2019joint, bruguier2019phoebe}, the confusion between similar phrases is observed and reduced by injecting phoneme information or training with difficult negative examples. A typical instance is the confusion between ``\texttt{Joan}" and ``\texttt{John}". The all-neural attention-based methods represent these two names with phrase-level embeddings and thus cannot effectively describe subtle differences at the token level. Likewise, these obstacles exist in the extensions of the attention-based method on other models, such as collaborative decoding (ColDec) \cite{han2021cif}.

ColDec transfers deep context to the CIF-based ASR \cite{cif} in a more controllable way. The CIF module in the CIF-based model non-uniformly compresses acoustic feature sequence along the time axis according to the acoustic boundaries of tokens, and emits token-level acoustic embeddings. These token-level acoustic embeddings provide a bridge to integrate textual context knowledge at the acoustic level. In ColDec, an extra context processing network (CPN) is trained to predict where to output which phrase according to the relevance between token-level acoustic embedding and contextual contents. Unlike deep context, ColDec combines the ASR outputs and the CPN biased outputs with a tunable weight to conduct collaborative decoding, thus leaving room for controlling in practice.

This paper improves the basic ColDec by dealing with the confusion with three techniques: fine-grained contextual knowledge selection (FineCoS), context purification, and position information. For FineCoS, we first restrict the range of phrase candidates according to their relevance to local acoustic embeddings, and then use fine-grained attention to extract token-level contextual representation from all tokens in these phrase candidates. In inference, we purify the phrase-level contextual representation by re-integrating the most relevant phrases to reduce context confusion. Besides, the influence of injecting position information on contextual modeling is also explored at different granularities. In previous works, \cite{chen2019joint, bruguier2019phoebe} focus on improving phrase encoder with extra phoneme inputs to discriminate similar phrases better, while our work mainly improves attention modules and the decoder side, and meanwhile fully use fine-grained knowledge. Compared with training with difficult negative examples \cite{alon2019contextual}, our work focuses more on neural contextual modeling instead of training strategies. 

\vspace{-5pt}

\section{Methods}
\label{sec:methods}

\subsection{Collaborative Decoding}
\label{ssec:coldec}

\vspace{-0.5pt}

\begin{figure}[t]
    \centering
    \includegraphics[width=\linewidth]{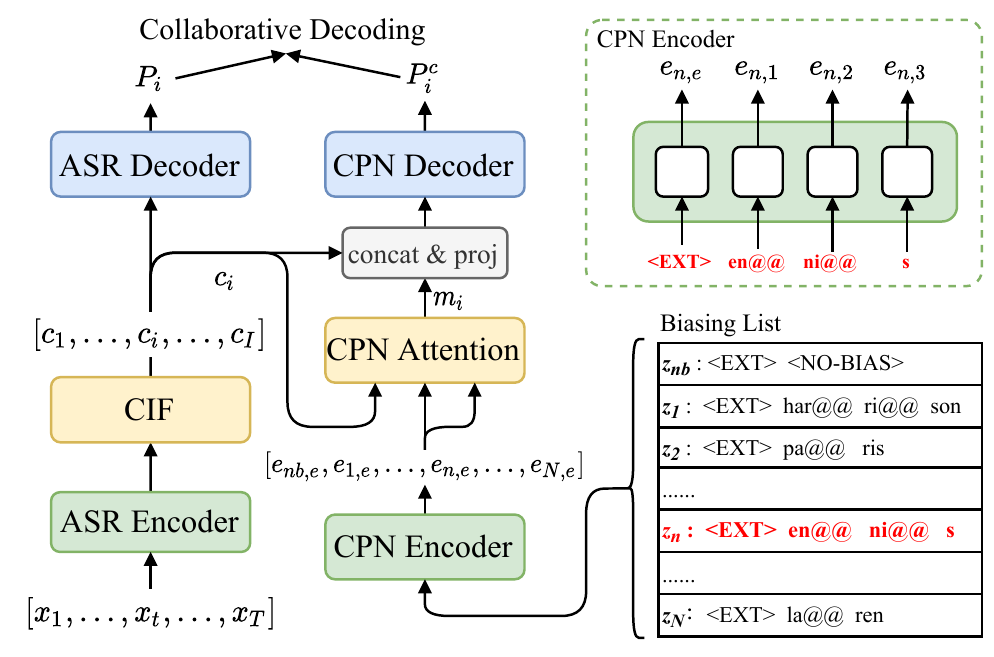}
    \vspace{-16pt}
    \caption{Collaborative Decoding on the CIF-based ASR: 1) the structure on the left is the CIF-based ASR model, while the structure on the right is context processing network (CPN); 2) At the top right corner shows how CPN encoder encodes a phrase and output its phrase embedding $e_{n,e}$ and all token embeddings $[e_{n,1}, e_{n,2}, e_{n,3}]$.}
    \label{ColDec}
    \vspace{-13.5pt}
\end{figure}

Collaborative decoding (ColDec) \cite{han2021cif} introduces phrase-level contextual modeling and attention-based relevance modeling to contextualize the CIF-based ASR model. In ColDec, besides the ASR model, a context processing network (CPN) is trained to extract target biasing phrases from transcription. For example, given reference ``\texttt{en@@ ni@@ s tri@@ ed to sleep}" and a target phrase ``\texttt{en@@ ni@@ s}", we generate the training target of CPN ``\texttt{en@@ ni@@ s \# \# \# \#}" from the reference by keeping the tokens of target phrase and replacing other tokens with ``\texttt{\#}" (which represents no biasing output). As shown in Fig.\ref{ColDec}, the CIF-based model consists of an encoder, a CIF module and a decoder. And the CPN comprises an encoder, an attention module and a decoder. This method is named collaborative decoding because the token-level acoustic embedding sequence $[c_1, ..., c_i, ..., c_I]$ emitted by the CIF module drives the decoding of the ASR model and the CPN simultaneously. 

Specifically, given raw biasing phrases $[z_1, ..., z_n, ..., z_N]$, a no-bias option $z_{nb}$ (represented as token ``\texttt{$<$NO-BIAS$>$}") is introduced as an option of not using context, and then a token ``\texttt{$<$EXT$>$}" is added to the start of each phrase for phrase embedding extraction. An example of processed biasing list $Z=[z_{nb}, z_1, ..., z_n, ..., z_N]$ is shown at the bottom right corner of Fig.\ref{ColDec}. At the top right corner of Fig.\ref{ColDec}, the CPN encoder encodes $n$-th phrase $z_n$ into a fixed-dimensional phrase embedding denoted as $e_{n,e}$ ($e_{nb,e}$ for no-bias option). With all phrase embeddings $[e_{nb,e}, e_{1,e}, ..., e_{n,e}, ..., e_{N,e}]$ as keys/values, the CPN attention consumes an token-level acoustic query $c_i$ to generate phrase-level contextual representation $m_i$. Finally, $c_i$ and $m_i$ are concatenated and subsequently sent to CPN decoder. In inference, driven by the CIF output $c_i$, ASR decoder and CPN decoder conduct decoding with interpolated log probability ($logP_{i} + \lambda logP_{i}^c$, where $\lambda$ controls the degree of biasing). Note that for CPN, the biasing phrases of each training batch is randomly sampled from n-grams in references, while the biasing phrases of test sets are usually extracted manually from context.

\subsection{Improvements for Confusion Reduction}

\subsubsection{Fine-grained contextual knowledge selection}
\label{ssec:Fine-grained contextual knowledge selection}

\textbf{Fine}-grained \textbf{Co}ntextual knowledge \textbf{S}election (FineCoS) introduces token-level contextual knowledge to reduce the uncertainty of token predictions. At first, we apply elementwise addition on token embedding $e_{n,j}$ (where $j$ denotes the index of token and $n$ denotes the index of phrase) emitted by the CPN encoder and its corresponding phrase embedding $e_{n,e}$ to generate the final token embedding $\bar{e}_{n,j}$. This operation informs the model of which phrase a token embedding belongs to. Similar to phrase-level attention, a token-level no-bias option $\bar{e}_{nb,1}$ is introduced to represent not using contextual knowledge. After the addition operation, token-level attention, which can be seen as ``soft selection", captures the relevance between acoustic embedding $c_i$ and final token embeddings $[\bar{e}_{nb,1}, \bar{e}_{1,1}, ..., \bar{e}_{n,1}, \bar{e}_{n,2}, ..., \bar{e}_{N,1}, \bar{e}_{N,2}, ...]$, and outputs the token-level contextual representation $g_i$. At last, the concatenation of $g_i$ and the output state of the CPN decoder is passed through a projection layer followed by a softmax layer to predict CPN targets.

\begin{figure}[t]
    \centering
    \includegraphics[width=\linewidth]{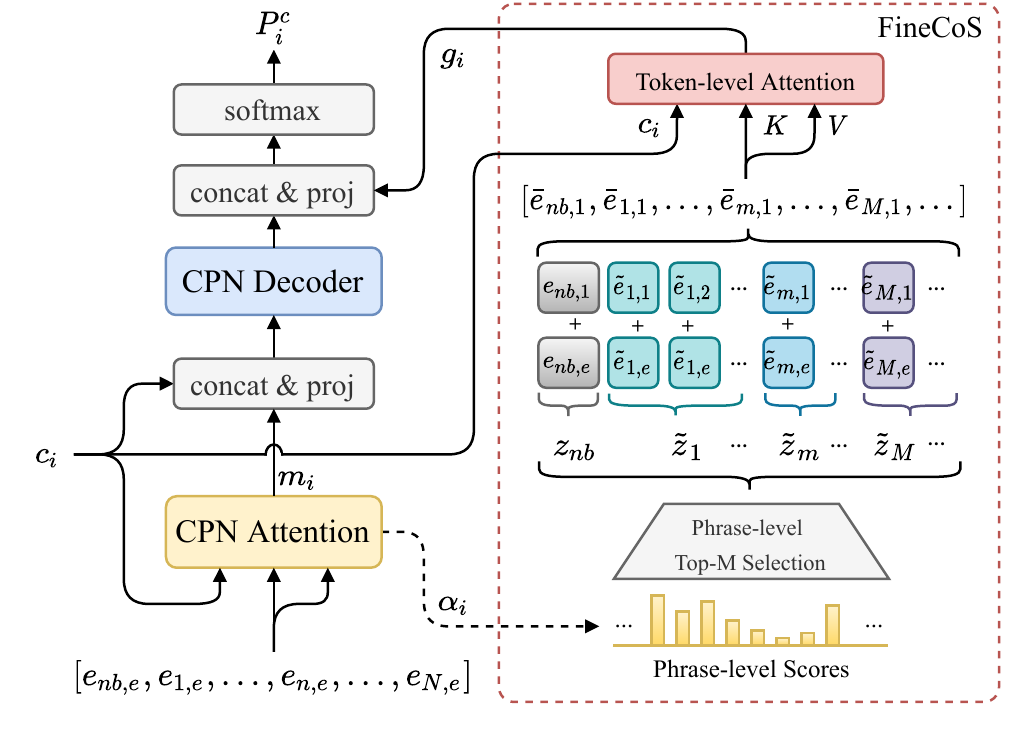}
    \vspace{-16pt}
    \caption{Fine-grained contextual knowledge selection. }
    \label{FineCoS}
    \vspace{-13.5pt}
\end{figure}

Unfortunately, the biasing phrases usually number in the thousands, which makes token-level attention intractable. Thus, we apply phrase-level hard selection to narrow the range of phrase candidates. Specifically, the top $M$ phrases are selected from $Z$ according to the ranking of CPN attention weight $\alpha_{i,n}$. After obtaining the top $M$ relevant phrases $[\tilde{z}_1, ..., \tilde{z}_m..., \tilde{z}_M]$ plus no-bias option $z_{nb}$, we retain the token embeddings from these $M$ phrases and conduct token-level attention with these retained token embeddings $[\bar{e}_{1,1}, ..., \bar{e}_{m,1}, ..., \bar{e}_{M,1}, ...]$ plus token-level no-bias option $\bar{e}_{nb,1}$ as keys/values. Here, phrase selection with the averaged CPN attention of the whole sequence  ($\check{\alpha}_{i,n} = {\frac{1}{I}}{\sum^{I}_{q=1}} {\alpha_{q,n}}$) is marked as ``global", while phrase selection with the averaged CPN attention of current step and past several steps  ($\check{\alpha}_{i,n} = {\frac{1}{Q}}{\sum^{i}_{q=i-Q+1}} {\alpha_{q,n}}$) is marked as ``local". The process of FineCoS shown in Fig.\ref{FineCoS} is written as 
\begin{align}
    &[z_{nb}, \tilde{z}_1, ..., \tilde{z}_m..., \tilde{z}_M] = \notag \\
    & \quad \quad \quad \quad \texttt{PhraseSelection}(Z, [\check{\alpha}_{i,1}, ..., \check{\alpha}_{i,n}, ..., \check{\alpha}_{i,N}]), \\
    &[\tilde{e}_{m,e}, \tilde{e}_{m,1}, ..., \tilde{e}_{m,j}, ...] = \texttt{CPNEnc}(\tilde{z}_m), \\
    &[{e}_{nb,e}, {e}_{nb,1}] = \texttt{CPNEnc}(z_{nb}), \\
    &\bar{e}_{m,j} = \tilde{e}_{m,j} + \tilde{e}_{m,e}, \quad\quad \bar{e}_{nb,1} = {e}_{nb,1} + {e}_{nb,e}, \\
    &K = V = [\bar{e}_{nb,1}, \bar{e}_{1,1}, ..., \bar{e}_{m,1}, ..., \bar{e}_{M,1}, ...]^T, \\
    &g_i = \texttt{TokenAttention}(c_{i}^T, K, V).
\end{align}

\subsubsection{Context purification}
\label{sec:context purification}

Context purification is proposed to make phrase-level contextual representation $m_i$ focus more on relevant contextual knowledge during inference. At each time step $i$, the most relevant phrases take up a large proportion of attention, while the rest only provide very limited information for phrase-level contextual modeling. Thus, discarding these redundant phrases makes $m_i$ focus more on relevant biasing phrases. Specifically, the top $K$ biasing phrases are selected according to the ranking of their CPN attention weights. Then, their corresponding attention weights $[\tilde{\alpha}_{i,1}, ..., \tilde{\alpha}_{i,k}, ..., \tilde{\alpha}_{i,K}]$ are re-normalized, which makes the sum of this partial distribution equal to $1$. Finally, the selected $K$ phrase embeddings $[\tilde{e}_{1,e}, ..., \tilde{e}_{k,e}, ..., \tilde{e}_{K,e}]$ and their re-normalized attention weights $[\hat{\alpha}_{i,1}, ..., \hat{\alpha}_{i,k}, ...,\hat{\alpha}_{i,K}]$ are combined via weighted sum as the purified contextual representation $m_i$. Context purification is applied at each time step in inference, and its procedure is roughly written as 
\begin{align}
    &[(\tilde{\alpha}_{i,1}, \tilde{e}_{1,e}), ..., (\tilde{\alpha}_{i,k}, \tilde{e}_{k,e}), ..., (\tilde{\alpha}_{i,K}, \tilde{e}_{K,e})] \notag \\
    & \quad \quad \quad \quad \quad \quad = \texttt{TopK}([\alpha_{i,nb}, \alpha_{i,1}, ...,\alpha_{i,n},..., \alpha_{i,N}]), \\
    &\hat{\alpha}_{i,k} = \frac{\tilde{\alpha}_{i,k}}{\sum^{K}_{k=1} {\tilde{\alpha}_{i,k}}}, \quad\quad m_i = {\sum^K_{k=1}} \hat{\alpha}_{i,k} \tilde{e}_{k,e},
\end{align}
where tilde denotes ``selected'' and hat denotes ``re-normalized''. Context purification is independently proposed in this work, but similar to the weak-attention suppression (WAS) \cite{shi2020weak} in some aspects. Both context purification and WAS mask some attention weights and re-normalize the rest, but context purification masks weights according to the ranking of weights and only functions in inference.

\subsubsection{Position information}
\label{sec:Position Information}

Compared with ColDec \cite{han2021cif}, the influence of position information on contextual modeling is explored at the phrase level and the token level. We inject position information to the inputs of the CPN encoder via position encoding \cite{vaswani2017attention}. Intuitively, the position encoding helps the CPN encoder model the differences in the token position distributions of phrases. Moreover, position information makes tokens more distinguishable (less confusing) because it tells the model which part of phrase a token is located in.

\section{Experimental Setup}
\label{sec:experiments}

\subsection{Datasets and Metrics}
\label{sssec:data and metrics}

Our experiments are conducted on LibriSpeech \cite{panayotov2015LibriSpeech} and an in-house code-switch ASR dataset. For LibriSpeech, we use 960 hours of labelled audio for training. As the test sets of LibriSpeech lack biasing lists, we construct biasing lists by collecting words that fall outside the 20,000 most common words in training data from references of test set and discarding short words that have less than 5 English letters. Finally, the simulated biasing lists for test-clean and test-other are composed of 1171 and 1129 phrases, respectively.

Our in-house dataset consists of $\sim$160,000 hours of labelled English and Chinese Mandarin audios collected from videos and other common acoustic situations. Its test sets are named test-name and test-term, both of which are collected from real in-house meetings. The details are shown in Table \ref{inhouse_testsets}. In the default biasing list, the total number of phrases of both biasing lists are 633 and 2415, and the number of distractors in them are 600 and 1775. Rare proportion, which denotes the proportion of rare non-distractor phrases (which appear less than 200 times in the training set) in all non-distractor phrases, is shown in Table \ref{inhouse_testsets}. Note that all biasing lists are session-level because we assume that audios are collected from meetings, so that all utterances in one test set share the same biasing list. We use word error rate (WER) for LibriSpeech and character error rate (CER) for the in-house dataset to measure ASR, and use precision (P), recall (R) and f1-score (F1) of biasing phrases to measure contextual biasing. More details about datasets are released\footnote{\url{https://github.com/MingLunHan/CIF-ColDec}}.

\begin{table}[t]
    \vspace{-15pt}
    \centering
    \caption{Details of in-house test sets and session-level biasing lists.}
    \vspace{4.5pt}
    \begin{tabular}{lp{2.2cm}<{\centering}p{2.2cm}<{\centering}}
    \toprule
    Details & test-name & test-term \\
    \midrule
    \# Biasing Utterances & 654 & 916 \\
    \# Total Utterances   & 748 & 1219 \\
    \# Distractors        & 600 & 1775 \\
    \# Total Phrases        & 633 & 2415 \\
    Rare Proportion (\%) & 78.79 & 39.53 \\
    Phrase Type & names & technical terms \\
    \bottomrule
    \end{tabular}%
    \label{inhouse_testsets}%
    \vspace{-12pt}
\end{table}%

\subsection{Configurations}
\label{ssec:configurations}

The input features are 80-dimension log-Mel filter banks extracted with 25ms window length and 10ms frame-shift via Kaldi \cite{Povey_ASRU2011}. For the in-house dataset, the frequency masking and time masking in SpecAugment \cite{park2019specaugment} are applied, and its ASR output vocabulary comprises 5740 Chinese characters and 4013 English word-pieces generated via BPE \cite{sennrich-etal-2016-neural}. For LibriSpeech, speed perturbation \cite{ko2015audio} with fixed $\pm$ 10\% and adaptive SpecAugment \cite{adaptivemaskingspecaug} are applied, and its ASR output vocabulary comprises 3726 word-pieces. For the CPN, its output vocabulary includes an extra token ``\texttt{\#}". 

The structure of the ASR model is almost the same as that in \cite{cif}. For LibriSpeech, we use 17 conformer blocks \cite{gulati2020conformer} as the ASR encoder, and 2-layer self-attention networks (SANs) as the ASR decoder. The hidden size $d_{model}$, the projection dimension $d_{ff}$ and the number of attention heads $h$ are 512, 2048 and 8, respectively. For the in-house dataset, we use 15-layer SANs as the ASR encoder, and 2-layer SANs as the ASR decoder ($d_{model}=640,d_{ff}=2560,h=8$). As for CPN, we use the same structure for both datasets. The CPN comprises an encoder with 4-layer SANs, a decoder with 2-layer SANs and an attention module with one SAN. The structure of proposed token attention is the same as that of CPN attention. The training of CPN is similar to that in \cite{han2021cif}. With the trained ASR model being frozen, we train the CPN with the same audios. Here, the probability of discarding sampled n-grams in training is 0.3, and the number of contextual batches in gradient accumulation \cite{han2021cif} is 8 for LibriSpeech and 2 for the in-house dataset. $M$ in FineCoS is 5, and $K$ in context purification is 2 for LibriSpeech and 10 for in-house dataset. In inference, we conduct search with beam size 10. More details about configurations of LibriSpeech are also released.

\section{Results}
\label{sec:results}

\subsection{Results on LibriSpeech}
\label{ssec:results on LibriSpeech}

\begin{table}[!hp]
    \vspace{-15pt}
    \centering
    \caption{Results on test-clean and test-other (\%).}
    \vspace{4.5pt}
    \begin{tabular}{p{3.58cm}p{0.62cm}<{\centering}p{0.62cm}<{\centering}p{0.62cm}<{\centering}p{0.62cm}<{\centering}}
    \toprule
    \multirow{2}[4]{*}{Method}  & \multicolumn{2}{c}{test-clean}    & \multicolumn{2}{c}{test-other} \\
    \cmidrule(r{3pt}){2-3} \cmidrule(lr){4-5}  & \multicolumn{1}{c}{F1}            & \multicolumn{1}{c}{WER} & \multicolumn{1}{c}{F1} & \multicolumn{1}{c}{WER} \\
    \midrule
    \texttt{S0}: CIF \cite{cif} & \multicolumn{1}{c}{85.22} & \multicolumn{1}{c}{2.12} & \multicolumn{1}{c}{71.89} & \multicolumn{1}{c}{5.26} \\
    \texttt{S1}: \texttt{S0} + {\footnotesize{ColDec}} \cite{han2021cif} & \multicolumn{1}{c}{85.79} & \multicolumn{1}{c}{2.14} & \multicolumn{1}{c}{73.36} & \multicolumn{1}{c}{5.29} \\
    \texttt{S2}: \texttt{S1} + {\footnotesize{Context Purification}} & \multicolumn{1}{c}{86.79} & \multicolumn{1}{c}{2.10} & \multicolumn{1}{c}{75.31} & \multicolumn{1}{c}{5.27} \\
    \texttt{S3}: \texttt{S2} + {\footnotesize{Position Information}} & \multicolumn{1}{c}{87.55} & \multicolumn{1}{c}{2.06} & \multicolumn{1}{c}{76.35} & \multicolumn{1}{c}{5.24} \\
    \texttt{S4}: \texttt{S3} + {\footnotesize{FineCoS (Global)}} & \multicolumn{1}{c}{88.82} & \multicolumn{1}{c}{2.01} & \multicolumn{1}{c}{77.72} & \multicolumn{1}{c}{5.15} \\
    \bottomrule
    \end{tabular}%
    \label{LibriSpeech_res}%
    \vspace{-6.5pt}
\end{table}%

We validate our improvements on LibriSpeech with the simulated biasing lists. As shown in Table \ref{LibriSpeech_res}, the CIF-based ASR (\texttt{S0}) achieves 2.12\% WER on test-clean and 5.26\% on test-other, which is a strong ASR baseline without extra language models. First, the basic ColDec is applied on both test sets. ColDec (\texttt{S1}) brings limited F1 improvement and slight WER degradation. We hypothesize the reason for the poor performance in \texttt{S1} is that \texttt{S0} has achieved rather good results on recognizing biasing phrases, and thus the recall of biasing phrases does not overwhelm the errors brought by over-biasing. Then, both context purification (\texttt{S2}) and position information (\texttt{S3}) bring F1 improvement and WER reduction on both test sets. Finally, the FineCoS (\texttt{S4}) further strengthens the ColDec and helps achieve 5.2\%/2.1\% relative WER reduction on test-clean/test-other, when compared to ASR baseline (\texttt{S0}).

\subsection{Results on In-house Large-scale Dataset}
\label{ssec:results on in-house dataset}

\begin{table*}[t]
    \vspace{-15pt}
    \centering
    \caption{Results on test-name and test-term: the enhanced ColDec is compared with the CIF-based ASR model (\texttt{E0}) and the CIF-based ASR model equipped with the basic ColDec (\texttt{E1}). In addition, the relative CER reduction compared with \texttt{E1} is provided.}
    \vspace{4.5pt}
    \begin{tabular}{p{4.3cm}p{0.92cm}<{\centering}p{0.92cm}<{\centering}p{0.92cm}<{\centering}p{2.1cm}p{0.92cm}<{\centering}p{0.92cm}<{\centering}p{0.92cm}<{\centering}p{2cm}}
        \toprule
        \multirow{2}[4]{*}{Method} & \multicolumn{4}{c}{test-name} & \multicolumn{4}{c}{test-term} \\
        \cmidrule(r{3pt}){2-5} \cmidrule(lr){6-9}          & P (\%) & R (\%) & F1 (\%) & CER (\%) & P (\%) & R (\%) & F1 (\%) & CER (\%) \\
        \midrule
        \texttt{E0}: CIF \cite{cif} & 100.00 & 27.23 & 42.81 & 11.28 & 93.15 & 71.98 & 81.21 & 16.69 \\
        \texttt{E1}: \texttt{E0} + ColDec \cite{han2021cif} & 99.79 & 68.01 & 80.92 & 9.16  & 88.97 & 82.34 & 85.52 & 16.45 \\
        \midrule
        \texttt{E2}: \texttt{E1} + Context Purification & 99.80 & 71.61 & 83.39 & 8.65 {\footnotesize($\downarrow$5.6\%)}  & 88.42 & 83.74 & 86.01 & 16.55 {\footnotesize($\uparrow$0.6\%)} \\
        \texttt{E3}: \texttt{E2} + Position Information & 99.81 & 73.54 & 84.68 & 7.93 {\footnotesize($\downarrow$13.4\%)}  & 87.01 & 84.28 & 85.62 & 16.40 {\footnotesize($\downarrow$0.3\%)} \\
        \midrule
        \texttt{E4}: \texttt{E1} + FineCoS (Global) & 99.44 & 76.37 & 86.39 & 8.64 {\footnotesize($\downarrow$5.7\%)} & 88.28 & 84.81 & 86.51 & 16.29 {\footnotesize($\downarrow$1.0\%)} \\
        \texttt{E5}: \texttt{E2} + FineCoS (Global) & 99.44 & 77.23 & 86.94 & 8.40 {\footnotesize($\downarrow$8.3\%)}  & 88.10 & 84.84 & 86.44 & 16.27 {\footnotesize($\downarrow$1.1\%)} \\
        \texttt{E6}: \texttt{E3} + FineCoS (Local,$Q=5$) & 99.81 & 76.51 & 86.62 & 7.94 {\footnotesize($\downarrow$13.3\%)}  & 85.55 & \textbf{87.84} & 86.68 & 16.33 {\footnotesize($\downarrow$0.7\%)} \\
        \texttt{E7}: \texttt{E3} + FineCoS (Global) & 99.82 & \textbf{77.95} & \textbf{87.54} & \textbf{7.66 {\footnotesize($\downarrow$16.4\%)}}  & 88.11 & 86.24 & \textbf{87.16} & \textbf{15.98 {\footnotesize($\downarrow$2.9\%)}} \\
        \bottomrule
    \end{tabular}%
    \label{main_results}%
    \vspace{-10pt}
\end{table*}%

\begin{table}[!htbp]
    \vspace{-16pt}
    \centering
    \caption{Comparison between ColDec with context purification (\texttt{E2}) and best enhanced ColDec (\texttt{E7}) with varying number of distractors.}
    \vspace{4.5pt}
    \begin{tabular}{p{1.7cm}p{1.2cm}<{\centering}p{1.2cm}<{\centering}p{1.2cm}<{\centering}p{1.2cm}<{\centering}}
        \toprule
        \multirow{2}[4]{*}{\# Distractors} & \multicolumn{2}{c}{\texttt{E2}} & \multicolumn{2}{c}{\texttt{E7}} \\
        \cmidrule(r{3pt}){2-3} \cmidrule(lr){4-5}          & F1 (\%) & CER (\%) & F1 (\%) & CER (\%) \\
        \midrule
        0     & 90.48 & 6.69  & 91.08 & 6.59 \\
        600 (default) & 83.39 & 8.65  & 87.54 & 7.66 \\
        1200  & 80.28 & 8.97  & 85.88 & 7.51 \\
        1800  & 76.84 & 9.02  & 85.31 & 7.77 \\
        2400  & 76.84 & 9.72  & 84.24 & 7.84 \\
        \bottomrule
    \end{tabular}%
    \label{comp_with_distractors}%
    \vspace{-6.5pt}
\end{table}%

Our in-house dataset is used to explore our methods on large-scale datasets and in real scenarios. As depicted in Table \ref{main_results}, compared with ASR baseline (\texttt{E0}), ColDec (\texttt{E1}) improves the recognition performance on both test sets. Then, we apply context purification and find that context purification (\texttt{E2}) brings both test sets with F1 improvements. Based on \texttt{E2}, \texttt{E3} injects position information to enhance contextual modeling, and shows F1 and CER improvements. Injecting position information brings 8.3\% relative CER reduction on test-name. To validate the effect of FineCoS, we build one branch (\texttt{E4}) from \texttt{E1} and two branches (\texttt{E6} and \texttt{E7}) from \texttt{E3}. Both \texttt{E6} and \texttt{E7} apply FineCoS, but \texttt{E6} uses FineCoS with local phrase selection ($Q=5$), and \texttt{E7} uses global phrase selection. Compared with \texttt{E3}, both \texttt{E6} and \texttt{E7} improve F1 on test sets, but only \texttt{E7} gets obvious CER reduction. To further validate the importance of position information on token-level contextual modeling, \texttt{E5} disables position information and introduces FineCoS. Though \texttt{E5} outperforms \texttt{E2}, it still falls behind \texttt{E7} if without position information, which proves that position information helps model fine-grained contextual knowledge. Our best model outperforms the basic ColDec with 6.65\% absolute F1 improvement and 16.4\% relative CER reduction on test-name, and with 1.64\% absolute F1 improvement and 2.9\% relative CER reduction on test-term. We conjecture that the improvements on test-name are due to the phrase-level confusion reduction brought by phrase selection and context purification, and the token-level uncertainty reduction brought by token attention that especially benefits rare word recognition.

As mentioned in \cite{pundak2018deep}, the large biasing list introduces high correlations between phrases, and thus causes performance degradation. In Table \ref{comp_with_distractors}, we investigate on test-name with \texttt{E2} and \texttt{E7} to verify the efficacy of position information and FineCoS on mitigating the degradation caused by the large biasing list. When not injecting distractors, the performance gain of \texttt{E7} over \texttt{E2} is limited. However, as the number of distractors grows, the gap between them widens. Finally, injecting 2,400 distractors causes 3.03\% CER degradation in \texttt{E2} but only causes 1.25\% CER degradation in \texttt{E7}. These results prove that position information and FineCoS cooperatively mitigate the degradation caused by large biasing lists.

\subsection{Further Analysis}
\label{ssec:further_analysis}

\begin{table}[!htbp]
    \vspace{-16pt}
    \centering
    \caption{Examples from left to right columns are references, hypotheses of \texttt{S1} and hypotheses of \texttt{S4}, respectively.}
    \vspace{4.5pt}
        \begin{tabular}{p{2.65cm}p{2.20cm}p{2.45cm}}
        \toprule
        the plays of marivaux (\footnotesize{\texttt{REF}}) & the plays of \textcolor{red}{marivox} (\footnotesize{\texttt{S1}}) & the plays of \textcolor{blue}{marivaux} (\footnotesize{\texttt{S4}}) \\
        \midrule
        sometimes as chiaroscurists (\footnotesize{\texttt{REF}}) & sometimes as \textcolor{red}{kioscurists} (\footnotesize{\texttt{S1}}) & sometimes as \textcolor{blue}{chiaroscurists} (\footnotesize{\texttt{S4}}) \\
        \bottomrule
        \end{tabular}%
    \label{cases}%
    \vspace{-6.5pt}
\end{table}%

To explore how FineCoS improves recognition, we extract examples from LibriSpeech test sets and analyze an example in terms of attention distribution. As shown in Table \ref{cases}, compared with \texttt{S1}, enhanced ColDec with FineCoS (\texttt{S4}) shows advantages on correcting token predictions. Here, we analyze the second example. For ``\texttt{ki@@}" in ``\texttt{kioscurists}" generated by \texttt{S1}, we found its top 3 most relevant phrase candidates are ``\texttt{chiaroscurist}", ``\texttt{chiaroscurists}", and no-bias option. This proves that the phrase-level CPN attention module captures the relevant contextual knowledge, but the CPN decoder does not fully use phrase-level contextual knowledge to correctly bias predictions. As for ``\texttt{chi@@}" in ``\texttt{chiaroscurists}" generated by \texttt{S4}, the selected phrases in FineCoS includes ``\texttt{chiaroscurist}" and ``\texttt{chiaroscurists}". As expected, the ``\texttt{chi@@}" in these two phrases dominates the token-level attention distribution (over 80\%), which validates that FineCoS narrows the range of phrase candidates and token candidates, and finally boosts token-level predictions.

\section{CONCLUSION}
\label{sec:conclusion}

In this paper, we improve E2E contextual speech recognition with fine-grained contextual knowledge selection, context purification, and position information, hoping these methods alleviate the confusion at different granularities. The proposed methods improve the basic contextual biasing method when studied on LibriSpeech and our large-scale in-house dataset. Although our method is developed on ColDec to customize the CIF-based ASR, we believe that our thoughts could be extended to other contextual biasing methods and other E2E models.

\vfill
\pagebreak

\bibliographystyle{IEEEbib}
\bibliography{strings,refs}

\begin{thebibliography}{10}

\bibitem{graves2006connectionist}
Alex Graves, Santiago Fern{\'{a}}ndez, Faustino~J. Gomez, and J{\"{u}}rgen
  Schmidhuber,
\newblock ``Connectionist temporal classification: labelling unsegmented
  sequence data with recurrent neural networks,''
\newblock in {\em {ICML}}. 2006, vol. 148 of {\em {ACM} International
  Conference Proceeding Series}, pp. 369--376, {ACM}.

\bibitem{graves2014towards}
Alex Graves and Navdeep Jaitly,
\newblock ``Towards end-to-end speech recognition with recurrent neural
  networks,''
\newblock in {\em {ICML}}. 2014, vol.~32 of {\em {JMLR} Workshop and Conference
  Proceedings}, pp. 1764--1772, JMLR.org.

\bibitem{graves2012sequence}
Alex Graves,
\newblock ``Sequence transduction with recurrent neural networks,''
\newblock {\em arXiv preprint arXiv:1211.3711}, 2012.

\bibitem{ChorowskiBSCB15}
Jan Chorowski, Dzmitry Bahdanau, Dmitriy Serdyuk, Kyunghyun Cho, and Yoshua
  Bengio,
\newblock ``Attention-based models for speech recognition,''
\newblock in {\em {NIPS}}, 2015, pp. 577--585.

\bibitem{chan2016listen}
William Chan, Navdeep Jaitly, Quoc~V. Le, and Oriol Vinyals,
\newblock ``Listen, attend and spell: {A} neural network for large vocabulary
  conversational speech recognition,''
\newblock in {\em {ICASSP}}. 2016, pp. 4960--4964, {IEEE}.

\bibitem{DBLP:conf/icassp/BahdanauCSBB16}
Dzmitry Bahdanau, Jan Chorowski, Dmitriy Serdyuk, Philemon Brakel, and Yoshua
  Bengio,
\newblock ``End-to-end attention-based large vocabulary speech recognition,''
\newblock in {\em {ICASSP}}. 2016, pp. 4945--4949, {IEEE}.

\bibitem{dong2018speech}
Linhao Dong, Shuang Xu, and Bo~Xu,
\newblock ``Speech-transformer: {A} no-recurrence sequence-to-sequence model
  for speech recognition,''
\newblock in {\em {ICASSP}}. 2018, pp. 5884--5888, {IEEE}.

\bibitem{aleksic2015bringing}
Petar~S. Aleksic, Mohammadreza Ghodsi, Assaf~Hurwitz Michaely, Cyril Allauzen,
  Keith~B. Hall, Brian Roark, David Rybach, and Pedro~J. Moreno,
\newblock ``Bringing contextual information to google speech recognition,''
\newblock in {\em {INTERSPEECH}}. 2015, pp. 468--472, {ISCA}.

\bibitem{hall2015composition}
Keith~B. Hall, Eunjoon Cho, Cyril Allauzen, Fran{\c{c}}oise Beaufays, Noah
  Coccaro, Kaisuke Nakajima, Michael Riley, Brian Roark, David Rybach, and
  Linda Zhang,
\newblock ``Composition-based on-the-fly rescoring for salient n-gram
  biasing,''
\newblock in {\em {INTERSPEECH}}. 2015, pp. 1418--1422, {ISCA}.

\bibitem{williams2018contextual}
Ian Williams, Anjuli Kannan, Petar~S. Aleksic, David Rybach, and Tara~N.
  Sainath,
\newblock ``Contextual speech recognition in end-to-end neural network systems
  using beam search,''
\newblock in {\em {INTERSPEECH}}. 2018, pp. 2227--2231, {ISCA}.

\bibitem{HeSPMAZRKWPLBSL19}
Yanzhang He, Tara~N. Sainath, Rohit Prabhavalkar, et~al.,
\newblock ``Streaming end-to-end speech recognition for mobile devices,''
\newblock in {\em {ICASSP}}. 2019, pp. 6381--6385, {IEEE}.

\bibitem{zhao2019shallow}
Ding Zhao, Tara~N. Sainath, David Rybach, Pat Rondon, Deepti Bhatia, Bo~Li, and
  Ruoming Pang,
\newblock ``Shallow-fusion end-to-end contextual biasing,''
\newblock in {\em {INTERSPEECH}}. 2019, pp. 1418--1422, {ISCA}.

\bibitem{pundak2018deep}
Golan Pundak, Tara~N. Sainath, Rohit Prabhavalkar, Anjuli Kannan, and Ding
  Zhao,
\newblock ``Deep context: End-to-end contextual speech recognition,''
\newblock in {\em {SLT}}. 2018, pp. 418--425, {IEEE}.

\bibitem{alon2019contextual}
Uri Alon, Golan Pundak, and Tara~N. Sainath,
\newblock ``Contextual speech recognition with difficult negative training
  examples,''
\newblock in {\em {ICASSP}}. 2019, pp. 6440--6444, {IEEE}.

\bibitem{chen2019joint}
Zhehuai Chen, Mahaveer Jain, Yongqiang Wang, Michael~L. Seltzer, and Christian
  Fuegen,
\newblock ``Joint grapheme and phoneme embeddings for contextual end-to-end
  {ASR},''
\newblock in {\em {INTERSPEECH}}. 2019, pp. 3490--3494, {ISCA}.

\bibitem{bruguier2019phoebe}
Antoine Bruguier, Rohit Prabhavalkar, Golan Pundak, and Tara~N. Sainath,
\newblock ``Phoebe: Pronunciation-aware contextualization for end-to-end speech
  recognition,''
\newblock in {\em {ICASSP}}. 2019, pp. 6171--6175, {IEEE}.

\bibitem{le2021deep}
Duc Le, Gil Keren, Julian Chan, Jay Mahadeokar, Christian Fuegen, and
  Michael~L. Seltzer,
\newblock ``Deep shallow fusion for {RNN-T} personalization,''
\newblock in {\em {SLT}}. 2021, pp. 251--257, {IEEE}.

\bibitem{le2021contextualized}
Duc Le, Mahaveer Jain, Gil Keren, et~al.,
\newblock ``Contextualized streaming end-to-end speech recognition with
  trie-based deep biasing and shallow fusion,''
\newblock {\em arXiv preprint arXiv:2104.02194}, 2021.

\bibitem{jain2020contextual}
Mahaveer Jain, Gil Keren, Jay Mahadeokar, Geoffrey Zweig, Florian Metze, and
  Yatharth Saraf,
\newblock ``Contextual {RNN-T} for open domain {ASR},''
\newblock in {\em {INTERSPEECH}}. 2020, pp. 11--15, {ISCA}.

\bibitem{han2021cif}
Minglun Han, Linhao Dong, Shiyu Zhou, and Bo~Xu,
\newblock ``Cif-based collaborative decoding for end-to-end contextual speech
  recognition,''
\newblock in {\em {ICASSP}}. 2021, pp. 6528--6532, {IEEE}.

\bibitem{cif}
Linhao Dong and Bo~Xu,
\newblock ``{CIF:} continuous integrate-and-fire for end-to-end speech
  recognition,''
\newblock in {\em {ICASSP}}. 2020, pp. 6079--6083, {IEEE}.

\bibitem{shi2020weak}
Yangyang Shi, Yongqiang Wang, Chunyang Wu, Christian Fuegen, Frank Zhang, Duc
  Le, Ching{-}Feng Yeh, and Michael~L. Seltzer,
\newblock ``Weak-attention suppression for transformer based speech
  recognition,''
\newblock in {\em {INTERSPEECH}}. 2020, pp. 4996--5000, {ISCA}.

\bibitem{vaswani2017attention}
Ashish Vaswani, Noam Shazeer, Niki Parmar, Jakob Uszkoreit, Llion Jones,
  Aidan~N. Gomez, Lukasz Kaiser, and Illia Polosukhin,
\newblock ``Attention is all you need,''
\newblock in {\em {NIPS}}, 2017, pp. 5998--6008.

\bibitem{panayotov2015LibriSpeech}
Vassil Panayotov, Guoguo Chen, Daniel Povey, and Sanjeev Khudanpur,
\newblock ``Librispeech: An {ASR} corpus based on public domain audio books,''
\newblock in {\em {ICASSP}}. 2015, pp. 5206--5210, {IEEE}.

\bibitem{Povey_ASRU2011}
Daniel Povey, Arnab Ghoshal, Gilles Boulianne, Lukas Burget, Ondrej Glembek,
  Nagendra Goel, Mirko Hannemann, Petr Motlicek, Yanmin Qian, Petr Schwarz,
  et~al.,
\newblock ``The kaldi speech recognition toolkit,''
\newblock in {\em IEEE 2011 workshop on automatic speech recognition and
  understanding}. IEEE Signal Processing Society, 2011.

\bibitem{park2019specaugment}
Daniel~S. Park, William Chan, Yu~Zhang, Chung{-}Cheng Chiu, Barret Zoph,
  Ekin~D. Cubuk, and Quoc~V. Le,
\newblock ``Specaugment: {A} simple data augmentation method for automatic
  speech recognition,''
\newblock in {\em {INTERSPEECH}}. 2019, pp. 2613--2617, {ISCA}.

\bibitem{sennrich-etal-2016-neural}
Rico Sennrich, Barry Haddow, and Alexandra Birch,
\newblock ``Neural machine translation of rare words with subword units,''
\newblock in {\em Proc. ACL}. 2016, ACL.

\bibitem{ko2015audio}
Tom Ko, Vijayaditya Peddinti, Daniel Povey, and Sanjeev Khudanpur,
\newblock ``Audio augmentation for speech recognition,''
\newblock in {\em {INTERSPEECH}}. 2015, pp. 3586--3589, {ISCA}.

\bibitem{adaptivemaskingspecaug}
Daniel~S. Park, Yu~Zhang, Chung{-}Cheng Chiu, Youzheng Chen, Bo~Li, William
  Chan, Quoc~V. Le, and Yonghui Wu,
\newblock ``Specaugment on large scale datasets,''
\newblock in {\em {ICASSP}}. 2020, pp. 6879--6883, {IEEE}.

\bibitem{gulati2020conformer}
Anmol Gulati, James Qin, Chung{-}Cheng Chiu, et~al.,
\newblock ``Conformer: Convolution-augmented transformer for speech
  recognition,''
\newblock in {\em {INTERSPEECH}}. 2020, pp. 5036--5040, {ISCA}.

\end{thebibliography}

\end{document}